\documentclass[journal=jctcce,manuscript=article,layout=twocolumn]{achemso}
\usepackage{amsfonts}
\usepackage{amsmath}
\usepackage{hyperref}
\usepackage{url}
\usepackage{float}
\usepackage[utf8x]{inputenc}
\usepackage{graphicx}
\usepackage{subfig}
\usepackage{epstopdf}
\usepackage{color}
\usepackage{textgreek}

\usepackage{xr}
\newcommand{\norm}[1]{\left\lVert#1\right\rVert}

\title{}

\author{S. Doerr}
\affiliation{Computational Biophysics Laboratory (GRIB-IMIM), Universitat Pompeu Fabra, Barcelona Biomedical Research Park (PRBB), C/ Doctor Aiguader 88, 08003 Barcelona, Spain}
\altaffiliation{Contributed equally to this work}

\author{I. Ariz-Extreme}
\affiliation{Computational Biophysics Laboratory (GRIB-IMIM), Universitat Pompeu Fabra, Barcelona Biomedical Research Park (PRBB), C/ Doctor Aiguader 88, 08003 Barcelona, Spain}
\altaffiliation{Contributed equally to this work}

\author{M. J. Harvey}
\affiliation{Acellera, Barcelona Biomedical Research Park (PRBB), C/ Doctor Aiguader 88, 08003 Barcelona, Spain}

\author{G. De Fabritiis}
\email{gianni.defabritiis@upf.edu}
\affiliation{Institució Catalana de Recerca i Estudis Avançats (ICREA), Passeig Lluis Companys 23, Barcelona 08010, Spain}
\affiliation{Computational Biophysics Laboratory (GRIB-IMIM), Universitat Pompeu Fabra, Barcelona Biomedical Research Park (PRBB), C/ Doctor Aiguader 88, 08003 Barcelona, Spain}

\title{Dimensionality reduction methods for molecular simulations}

\begin{document}

\begin{abstract}
Molecular simulations produce very high-dimensional data-sets with millions of data points. As analysis methods are often unable to cope with so many dimensions, it is common to
use dimensionality reduction and clustering methods to reach a reduced representation of the data. Yet these methods often fail to capture the most important features necessary for the construction of a Markov model. Here we demonstrate the results of various dimensionality reduction methods on two simulation data-sets, one of protein folding and another of protein-ligand binding. The methods tested include a $k$-means clustering variant, a non-linear auto encoder, principal component analysis and tICA. The dimension-reduced data is then used to estimate the implied timescales of the slowest process by a Markov state model analysis to assess the quality of the projection. The projected dimensions learned from the data are visualized to demonstrate which conformations the various methods choose to represent the molecular process.
\end{abstract}
\maketitle

\section{Introduction}
Molecular dynamics (MD) simulations allow one to simulate bio-molecules with increasingly good accuracy and in recent years have begun to provide meaningful predictions of 
experiments and insight into atomistic mechanisms, like the process of protein folding into native structures \cite{lindorff-larsen_how_2011}. From the computational point of view, one of the primary challenges of MD simulations is the ability to sample experimentally relevant millisecond to second timescales. With the advent of general-purpose graphics processing units in 2009\cite{harvey_acemd:_2009}, it has become possible to produce microseconds, and more recently milliseconds, of aggregated simulation data. This data is high dimensional with a common system size being of the order of ten to hundred thousand dimensions. The results are often analyzed using Markov state models (MSMs)\cite{prinz_markov_2011}. Discrete Markov state models require the definition of discrete states which are usually computed by clustering over a metric space. Depending on the metric used and the dimensionality of the space, the clustering might produce a poor discretization of states, hiding the slow dynamics and yielding a poor MSM from which it is impossible to compute the correct thermodynamic variables\cite{prinz_markov_2011}. As a consequence, it is important to use a proper metric space for each system and a proper discretization, i.e. one that captures the most relevant information about the simulated molecular process.

Choosing the most favorable reduced metric space for a system is difficult without \textit{a priori} information, and clustering over high dimensional spaces can be very 
challenging \cite{kriegel_clustering_2009}. In recent years, new algorithms that can learn complex functions have lead to methods which produce a lower dimensional
representation of the data that have no significant loss of information \cite{wang_folded_2012}. Sparse coding \cite{olshausen_emergence_1996}, auto encoders \cite{hinton_reducing_2006} and neighborhood embedding \cite{van_der_maaten_visualizing_2008} have shown to be very effective in reducing the dimensionality of data while preserving important underlying features. Dimensionality reduction methods have also been developed specifically for molecular dynamics data by reweighing features with unsupervised methods \cite{blochliger_weighted_2015}, by learning distance functions \cite{mcgibbon_learning_2013} and by using diffusion maps \cite{boninsegna_investigating_2015}.


In this work we focus on comparing the performance of dimensionality reduction methods on biological simulation data. We resolve the folding of a protein and the binding of a ligand to a protein by simulation and try to find the projection that produces the best MSM using non-linear auto encoders, clustering and linear projection methods such as PCA and tICA\cite{perez-hernandez_identification_2013}.

\section{Methods}
\subsection{Data-sets}

\begin{figure}[htp]
\subfloat[Villin]{
  \includegraphics[width=.4\columnwidth]{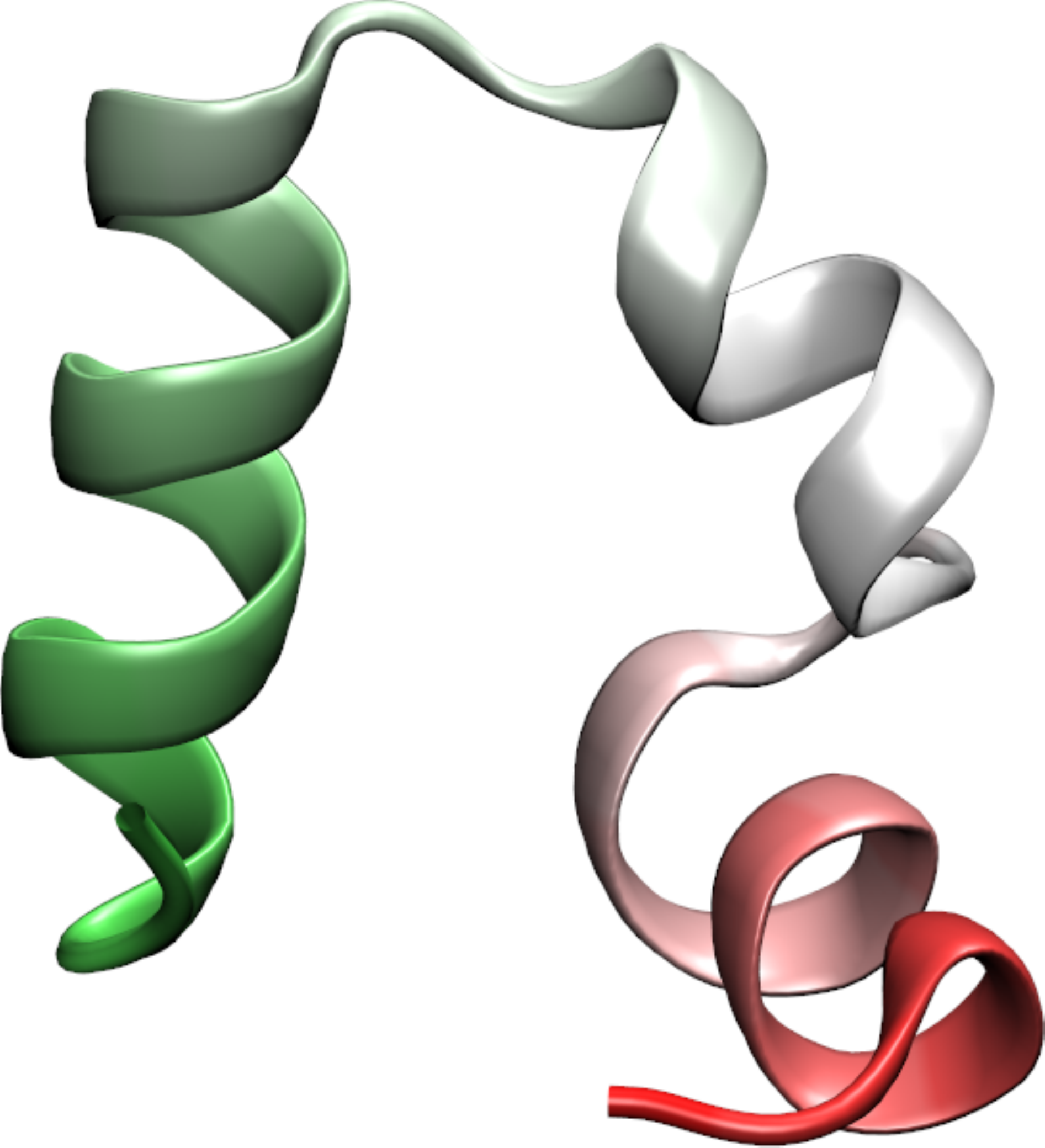}
}
\subfloat[Benzamidine-Trypsin]{
  \includegraphics[width=.6\columnwidth]{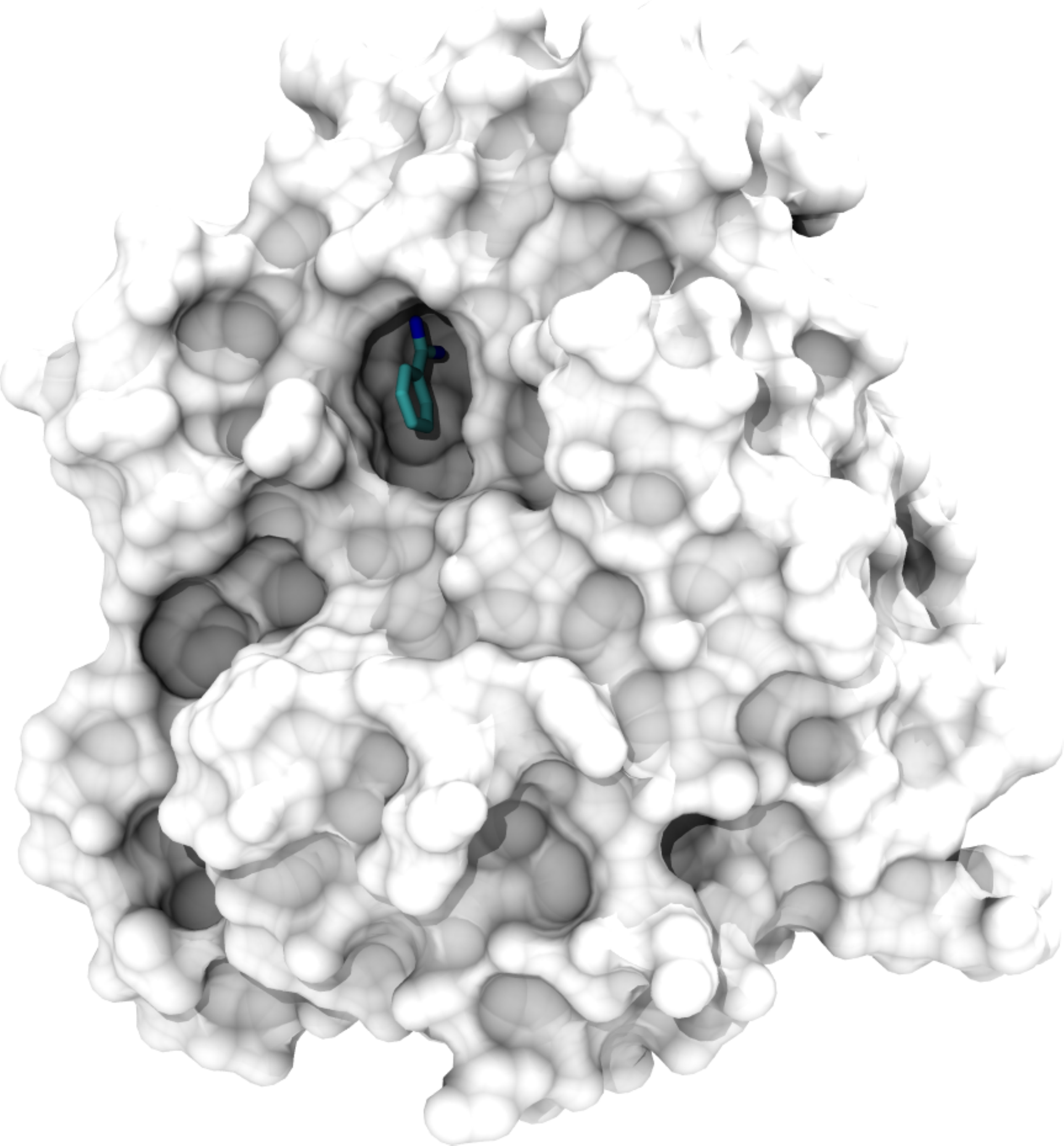}
}
\caption{a) folded structure of Villin. b) bound configuration of Benzamidine-Trypsin.}
\label{fig:folded}
\end{figure}

The data-sets used are from the folding and unfolding simulations of Villin as well as the ligand-binding simulations of Benzamidine to Trypsin. 

Villin (see folded structure in Fig. \ref{fig:folded}a) is a tissue-specific protein which binds to actin. The part under study is a double norleucin mutant of the $ 35 $ amino acid long headpiece widely tested in MD simulations because of its fast folding properties. At the temperature of $ 300^{\circ}K $ the non mutated protein domain has an experimental folding time of $ 4.3 \mu s $ \cite{kubelka_experimental_2003,kubelka_sub-microsecond_2006}. Computational estimations of the double mutant at $ 360^{\circ}K $ gave a folding time of $ 3.2 \mu s$ \cite {piana_protein_2012}, a folding free energy of $-0.6$ kcal/mol and a timescale of the order of $200ns$ and will be used as reference, as the same setup will be used here.

Benzamidine-Trypsin is a protein-ligand binding system, with an experimental free energy of $-6.2 kcal/mol$ \cite{mares-guia_studies_1965} at $ 300^{\circ}K $ and a  timescale of the binding process of the order of $600 ns$. 


The structure of Villin was taken from Piana et. al.\cite{piana_protein_2012}, solvated in water and simulated using the CHARMM22* forcefield \cite{lindorff-larsen_systematic_2012}. The Benzamidin-Trypsin setup was taken from Buch et. al.\cite{buch_complete_2011}, solvated and simulated using the AMBER 99SB force field \cite{hornak_comparison_2006}. Simulations were performed using ACEMD\cite{harvey_acemd:_2009}, a molecular dynamics code for graphical processing units, on the GPUGRID distributed computing infrastructure\cite{buch_high-throughput_2010}.

For Villin, 1562 simulations were used, each $120 ns$ long, resulting in an aggregate simulation time of $187.4 \mu s$ and 1,874,400 conformations at a sampling time of $0.1 ns$. For Benzamidine-Trypsin, 488 simulations of $100 ns$ were used for a total aggregate simulation time of $48.8 \mu s$ and 488.000 configurations. To best demonstrate the performance of the dimensionality reduction methods in scarce-data regimes which are the norm in MD simulations, we bootstrapped the data-sets 20 times at various percentages of the total data-set, thus obtaining various subsampled data-sets at 20-100\% of the total simulation data.

\subsection{Preprocessing of the data}
During simulations, the configuration of the system is represented by the positions and velocities of all atoms. For analysis purposes, however, a translation and rotation invariant representation is ideal. Therefore, for Villin we calculate and use the protein contact maps of the conformations and for Benzamidine-Trypsin we use the ligand-protein contact maps. 

For Villin, the contact maps were produced from the original trajectories by calculating the distance between the backbone $ C_{\alpha} $ of each amino acid to the $ C_{\alpha} $ atoms of all other amino-acids. Each element of the resulting distance matrix was transformed into 1 if the distance was below 8\AA\ and $0$ otherwise. As contact maps are symmetric, only the upper triangular part of the matrix was considered. The upper triangular part was then expanded into a vector of $\frac{n_{res} (n_{res}-1)}{2}$ contacts. For Villin with 35 residues, this results in 595-element binary-valued contact maps. The contact map data-set of Villin is on average 80\% sparse and fewer than 1\% of the contact maps are duplicates.

For Benzamidine-Trypsin, the protein-ligand contact maps were produced by calculating the distance between the $ C_{\alpha} $ atoms of the residues of Trypsin and two carbon atoms at opposite sides of the Benzamidine (as show in the SI of Doerr et al. \cite{doerr_--fly_2014}). The distances were then thresholded similarly to 8\AA\ as in Villin to produce contacts, however in this case, the contacts are a one-dimensional vector of $446$ contacts (2 ligand atoms times 223 protein residues). The contact maps of Benzamidin-Trypsin are on average 99\% sparse.

\subsection{Dimensionality reduction methods}
The data-sets have proven challenging to analyze using standard clustering methods like $k$-means, $k$-centers, and others. In particular the folded state of Villin is not easily detected and therefore an MSM built on top of such clustering would lose any information on the folding process. A cause could be the high dimensionality of the data which can spread out clusters which exist on subspaces. A projection of the data on a lower dimensional space can lead to an improvement in the clustering and MSM constructed on top of it. In this work, four different methods are used for learning the features of a lower dimensional representation: a modification of $k$-means \cite{coates_analysis_2011}, principal component analysis (PCA), a non-linear auto encoder and tICA\cite{perez-hernandez_identification_2013}. The motivation for this choice is that $k$-means is an unsupervised method commonly used for clustering bio-molecular data; PCA is an optimal projection method in the linear regime; tICA, another linear method, extends the idea of PCA by using the time component of the simulations and auto encoders are a good extension to the non-linear regime when a sigmoid function is used as the activation function. Auto encoders are also known to learn PCA under certain conditions \cite{bourlard_auto-association_1988}, i.e. linear activation function and transposed weights between encoding and decoding. A fifth method, called t-SNE\cite{van_der_maaten_visualizing_2008} was also considered due to its recent impressive success on various data-sets like the MNIST, NORB and NIPS as well as the Merck Viz Challenge. However, due to its high computational cost and memory requirements, we were not able to test it on our data-set.

\subsection{$k$-means triangle}
The $k$-means (triangle) method was taken from Coates et al. \cite{coates_analysis_2011}. Normal $k$-means clustering produces a hard assignment of each data point to a single cluster it corresponds to and can be represented by a binary $1 \times K$ vector (where K the number of clusters), which is 1 on the index of the closest cluster center and 0 elsewhere. $k$-means (triangle) on the other hand, after computing the cluster centers, represents each data point $x$ as a $1 \times K$ dimensional vector $v_x$ whose elements are calculated by $$v_x(i) = max\{0,\mu(z)-z_i\}$$ where $c^{(i)}$ is the $i$-th cluster center, $z_i = \|x-c^{(i)}\|$ is the distance of data point $x$ from cluster center $c^{(i)}$ and $\mu(z)$ is the mean of all $z_i$ of $x$. In other words, each data point gets represented by a vector of its distances to all cluster centers subtracted by their mean and thresholded at a minimum 0. This method proved superior in Coates et al. \cite{coates_analysis_2011} compared to normal $k$-means clustering and several other methods. In this study $k$-means (triangle) was used to project the contact map data into the $1 \times K$ space defined by the $K$ cluster centers.

\subsection{PCA}
Principal component analysis is one of the most widely used dimensionality reduction methods. By calculating the eigenvectors of the data covariance matrix, PCA can project the data on the principal components which are the dimensions of largest variance of the data-set, thereby minimizing the total squared reconstruction error of the projected data through a linear transformation of the input data. It enjoys wide application in the field of computational biology, implementations exist for most programming languages and it has a quick runtime. In this study we used PCA to project the contact map data onto the first principal components of PCA.

\subsection{tICA}
Time-lagged independent component analysis is a dimensionality reduction method recently rediscovered and applied very successfully to biological problems \cite{perez-hernandez_identification_2013,schwantes_improvements_2013}. The reason for its great success in such problems is that the tICA projections are the linear transform of
the input data which maximizes the auto-correlations of the output data. This means that it is able to identify and project the data on the slowest sub-space which can be obtained through a linear transform. As the biologically most interesting processes in simulations are often transitions between metastable states separated by large barriers, tICA is able to project the data onto those slow processes and thus allows a finer discretization of the slow dynamics without losing information related to those slow processes. In this study we used tICA to project the contact map data on the first time-lagged independent components.

\subsection{Auto encoder}
An auto encoder is a neural network which tries to reconstruct a given input vector in its output layer after encoding it in one or more hidden layers. Therefore, input and output layers of auto encoders have the same number of units while the hidden layers often contain fewer units than the input and output, thus forcing the neural network to learn a lower dimensional representation of the data. The activation of an auto encoder unit is defined by 
\begin{equation}\label{eq_features}
a_{i}^{(L+1)} = f\left(\sum_{j=1}^{S_{L}}W_{i j}^{(L)}a_{j}^{(L)} + b_{i}^{(L)}\right)
\end{equation}\\
where  $ a_{i}^{(L)} $ the $ i $-th unit in layer $ L $, $ S_{L} $ the number of units in layer $ L $, $W_{ij}^{(L)} $ the weight matrix of layer $L$, $ b_{j}^{(L)} $ the $ j $-th bias term and $ f $ is the activation function. Various activation functions can be used in an auto encoder, however as we as we want to test a non-linear auto encoder, we choose to use a \textit{sigmoid} function $f(z) = 1/(1 + e^{-z})$. Additionally a sigmoid output layer aids us in mapping the reconstructed data to contact maps as its output values are within the range $[0,1]$.

\begin{figure}[!tpb]
\centerline{\includegraphics[width=0.7\columnwidth]{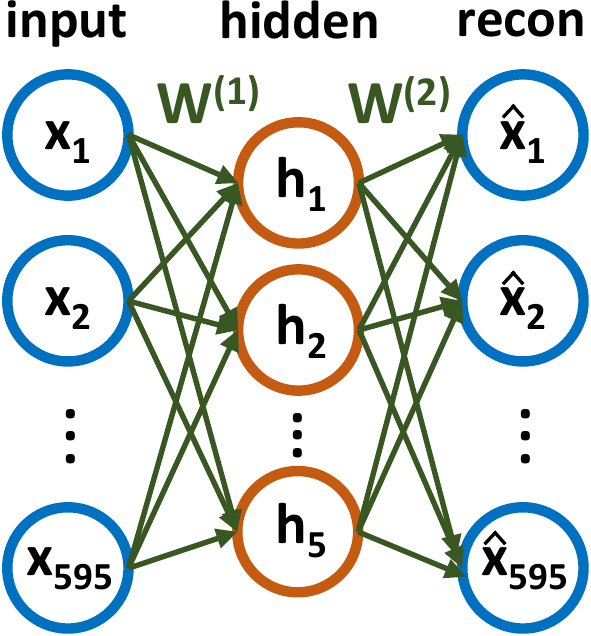}}
\caption{
Auto encoder architecture for a 5 dimensional projection of the Villin data. $x$ represents the input data in the input layer which consists of 595 units,
$\hat{x}$ represents the reconstructed output in the output layer with 595 units and h represents the data representation in 
the hidden layer consisting of various number of hidden units depending on the auto encoder. $W$ denotes the weights
applied to the activation of the layers before them.
}
\label{fig:autoencoder} 
\end{figure}

The most common optimization algorithm for auto encoders is gradient descent through the back propagation algorithm. Nevertheless, more elaborate algorithms have
been used such as the conjugate gradient, or the Hessian-free algorithm used by \cite{martens_deep_2010}, which is a $ 2^{nd} $-order optimization algorithm. Among these, the \textit{L-BFGS} algorithm explained by \cite{liu_limited_1989} has been shown by \cite{le_optimization_2011} to be among the most efficient ones and was used here. For the purposes of this study, we built various shallow (single hidden layer) auto encoders in Theano\cite{bergstra_theano:_2010}. The configuration for the 5-dimensional auto encoder can be seen in Figure \ref{fig:autoencoder}. After forward propagating the examples through the auto encoder, a cost function, given in equation \eqref{eq:ae}, is evaluated and the gradients for each layer are calculated by back-propagation.

The \textit{L-BFGS} algorithm is then used for training over 400 epochs. After training, the projected simulation data is obtained by removing the output layer of
the auto encoder and taking the lower dimensional representation produced by the hidden layer for each simulation frame.

\subsection{Markov state model}
MSMs have been  used to reconstruct equilibrium and kinetic properties in many molecular systems \cite{buch_complete_2011,doerr_--fly_2014,bisignano_kinetic_2014}. MSMs allow to extrapolate equilibrium properties of a dynamical system like MD, by many out-of-equilibrium trajectories. The trajectories have first to be discretized by assigning each frame to a given state. In this study, the projected data frames of each projection method were clustered and assigned to the closest of 1000 states produced by the mini-batch $k$-means algorithm of Scikit-learn \cite{pedregosa_scikit-learn:_2011}, thus producing discretized trajectories.

Using the discretized trajectories, a master equation can then be constructed by determining the frequency of transitions between states,
\begin{equation}\label{eq:ME}
\frac{dP_i(t)}{dt}=\sum_{j=1}^N \left[k_{ij} P_{j}(t)-k_{ji}P_i(t)\right] 
\end{equation}
where $P_i(t)$ is the probability of state i at time t, and $k_{ij}$ are the transition rates from j to i. The master equation (Eq. \ref{eq:ME}) can be rewritten in a compact matrix form $d\mathbf{P}/dt=\mathbf{K} \mathbf{P}$ where  $K_{ij} = k_{ij}$ for $i \neq j$ and $K_{ii}=-\sum_{j \neq i} k_{ji}$. The formal solution is $\mathbf{P}(t) = \mathbf{T} \mathbf{P}(0)$ where $\mathbf{T} = p(i,t|j,0)$ is the probability of being in state $i$ at time $t$, given that the system was in state $j$ at time 0. The transition matrix $\mathbf{T}$ is estimated from the simulation trajectory by counting how many transitions are observed between $i$ and $j$ and vice-versa and using a reversible maximum-probability estimator\cite{prinz_markov_2011}. 

From the the matrix $T$ all the thermodynamics and kinetics properties of the system can be determined as well as a kinetic lumping of clusters using the PCCA+ method\cite{cordes_metastable_2002}. 
The implied timescale of the slowest process, which we will focus on in this study, can be calculated from the second eigenvalue of the transition probability matrix $T$ as $t = \frac{\tau}{ln(\lambda(\tau))}$, where $t$ the slowest timescale, $\tau$ is the lag-time at which the Markov model is constructed and $\lambda(\tau)$ the second eigenvalue of the transition probability matrix of the Markov model.

\section{Results}
We projected the Villin and Benzamidine-Trypsin data-sets with the four dimensionality reduction methods and analyzed the projected data using Markov models. To demonstrate the performance of the methods under scarce data conditions we calculated Markov models containing decreasing amounts of simulations and to reduce the effect of individual trajectories on the result, we bootstrapped the simulations used in the model over 20 iterations. From the 20 bootstrapped Markov models we calculate the slowest implied timescale over a range of lag-times. A converged Markov model should have the slowest timescale converged over a range of lag-times and thus there should be a low standard deviation to the timescales calculated. Convergence of timescales is important for Markov models as it is an indication of Markovinianity of the model and is required for calculating consistent eigenvalues and eigenvectors and thus all other observables.

\subsection{Benzamidine-Trypsin}
\begin{figure}[htp]
\includegraphics[width=\columnwidth]{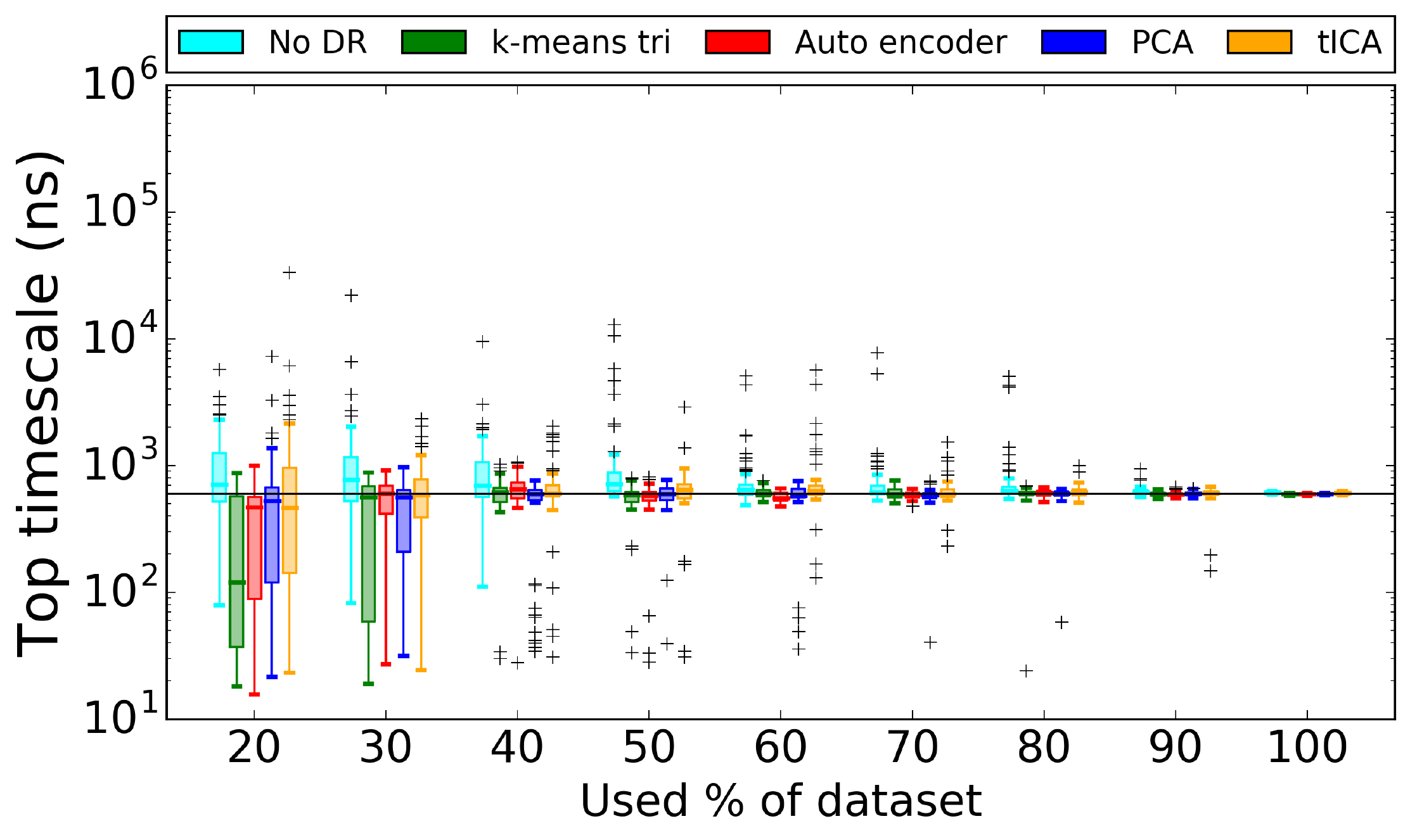}
\caption{Top implied timescales for Markov models built for Benzamidine-Trypsin using 5 dimensional projections. Timescales were estimated at lag-times of $5$ to $15ns$. The black horizontal line indicates the reference timescale of $600ns$.}
\label{fig:itsbentryp}
\end{figure}

In Figure \ref{fig:itsbentryp} we can see the performance of the dimensionality reduction methods reflected in the implied timescales of the Benzamidine-Trypsin data-set. We can see that on this data-set, even without dimensionality reduction we are able to obtain the correct timescale, with all methods showing very small errors, when using more than 50\% of the dataset. We should note however, that increasing the range of lag-times as in SI Figure \ref{fig:itsbentryplargelag} to $40ns$, we see that the full dimensional data, tICA and PCA, perform worse. In general this is not a big issue as Markov models are typically constructed at the shortest lag-time at which convergence is seen (in this case $10ns$). However, it shows us that at larger lag-times, the slow process can become lost in the three aforementioned methods. Interestingly, $k$-means (triangle) and the auto encoder are not affected by this and keep consistently converged timescales over large lag-times. Therefore, dimensionality reduction methods can help in this case with keeping the timescales flat over larger lag-times.

Changing the number of dimensions on the other hand does not have a significant effect on Benzamidine-Trypsin. Results are shown for 50 dimensions in SI Figure \ref{fig:itsbentryp50} without any notable changes, indicating that the dimensionality reduction methods at the very least do not produce a worse projection than the starting data.

\subsection{Villin}
\begin{figure*}[htp]
\subfloat[5 dimensions]{
  \includegraphics[width=.5\textwidth]{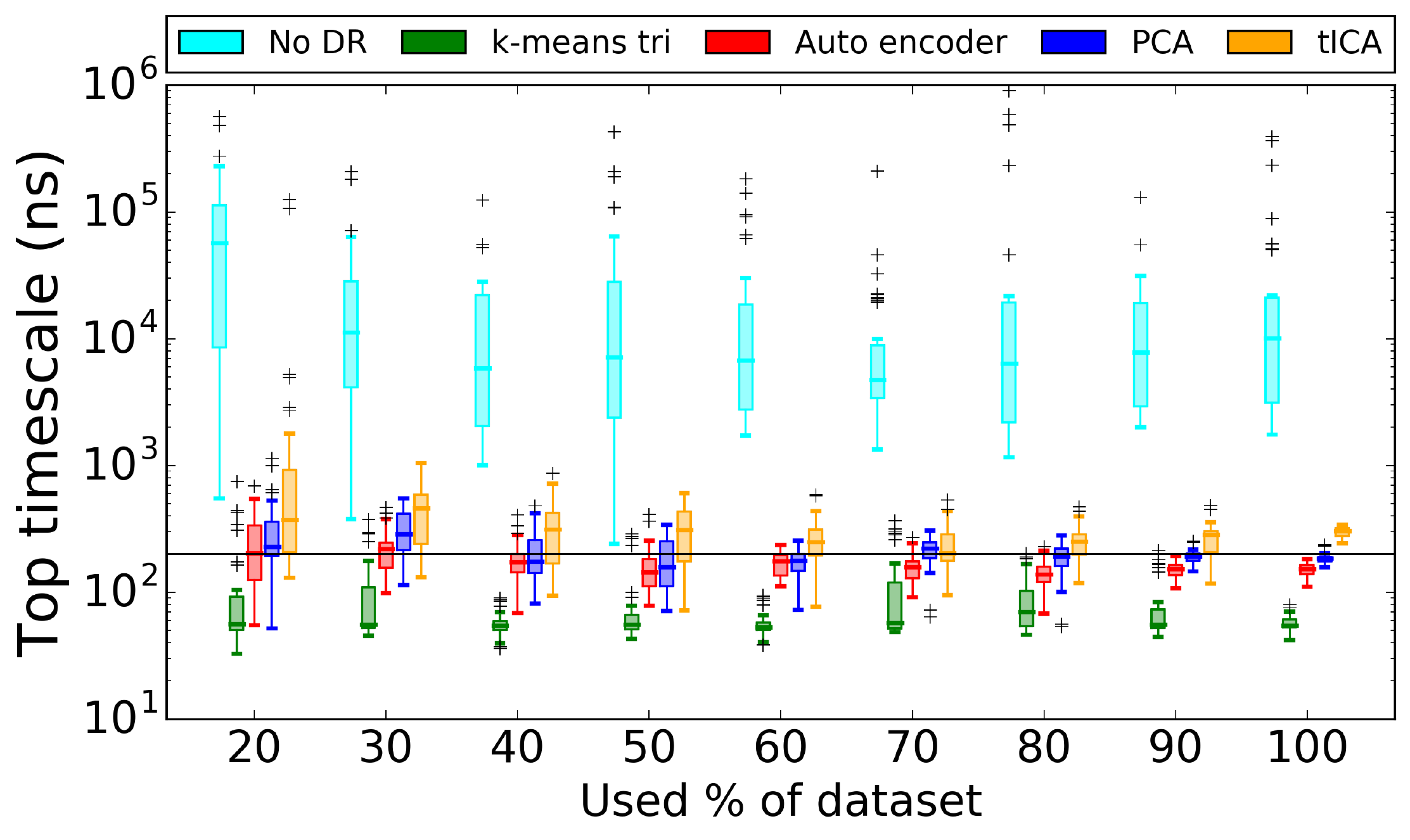}
}
\subfloat[20 dimensions]{
  \includegraphics[width=.5\textwidth]{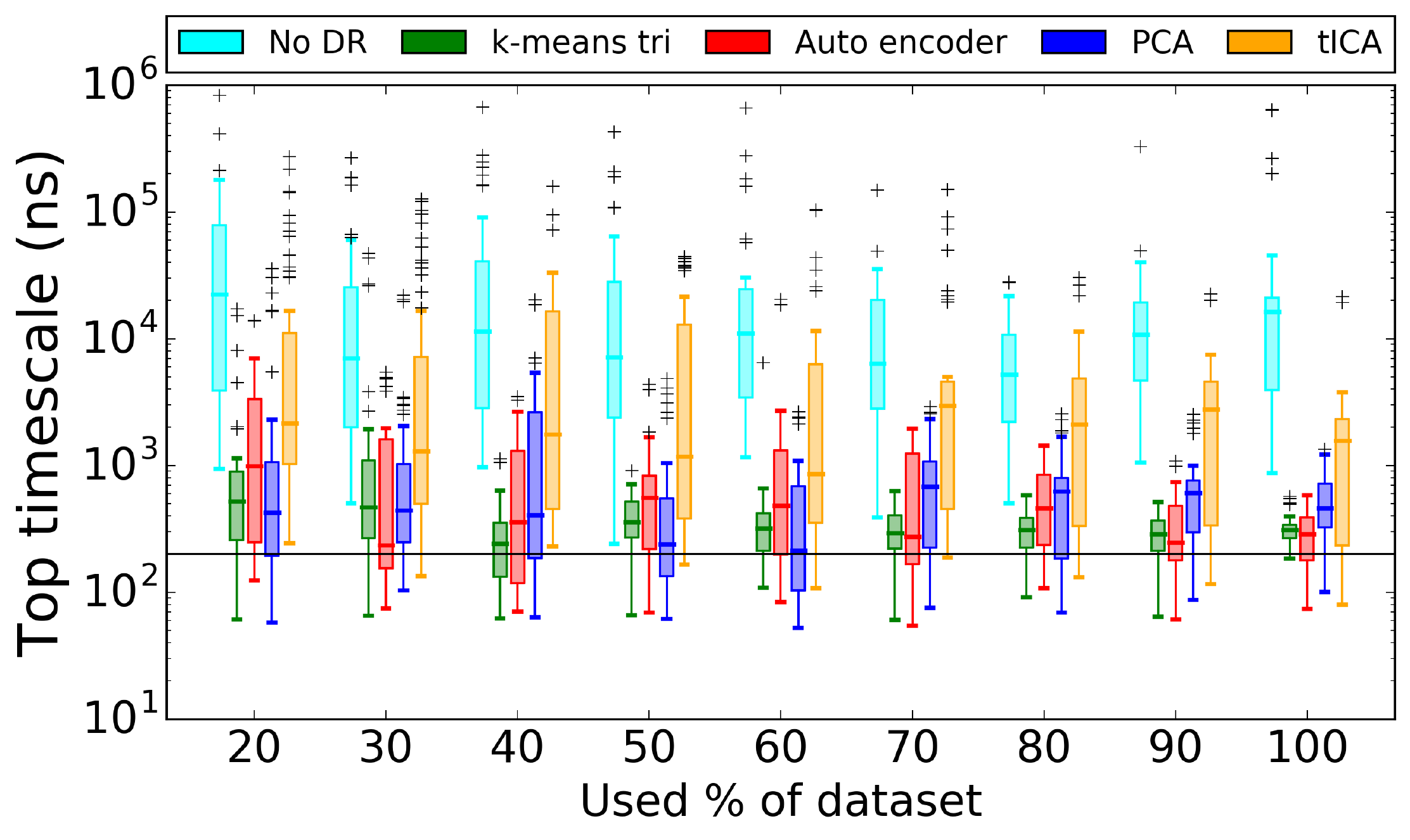}
}
\caption{Top implied timescales for Markov models built for Villin using 5 and 20 dimensional projections. Timescales were estimated at lag-times from $25$ to $30ns$. The black horizontal line indicates the reference timescale of $200ns$.}
\label{fig:itsvillin}
\end{figure*}

For Villin in Figure \ref{fig:itsvillin} it can be seen that using the full dimensional data is not an option as it overestimates the timescale by at least two orders of magnitude with huge uncertainty reflected in the error bars. This makes this system much more challenging than Benzamidine-Trypsin and stresses the importance of dimensionality reduction. In Figure \ref{fig:itsvillin}a which shows the 5-dimensional projections; out of the four projection methods, PCA, tICA and the auto encoder dominate, estimating the timescale for Villin closest to the reference timescale, with $k$-means (triangle) underestimating the timescale by a factor of around 3. However, by increasing the number of projected dimensions to 20 as in Figure \ref{fig:itsvillin}b, we see that with 20 dimensions the estimation errors of the timescales are much larger, with $k$-means (triangle), PCA and the auto encoder performing best.

\begin{figure}[htp]
\includegraphics[width=\columnwidth]{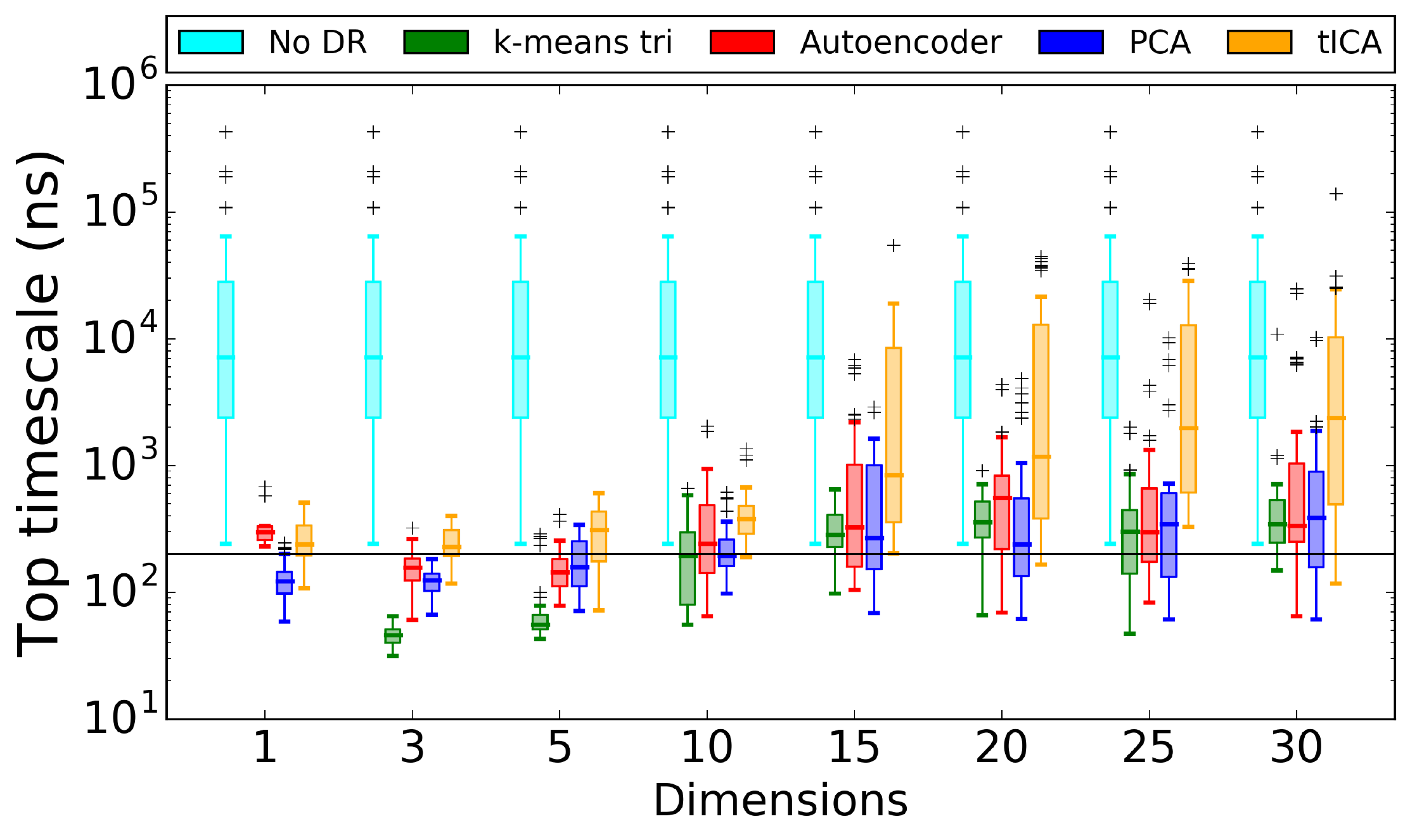}
\caption{Top implied timescales for Villin using a varying number of dimensions on the 50\% data-set. Timescales were estimated at lag-times of $25$ to $30ns$. The black horizontal line indicates the reference timescale of $200ns$.}
\label{fig:itsvillindim}
\end{figure}

Further investigating this, we tested at the 50\% data-set various numbers of dimensions. The results can be seen in Figure \ref{fig:itsvillindim}, which shows that for Villin the number of projected dimensions is critical for the construction of a working Markov model. The best performance is obtained by low-dimensional tICA and the auto encoder, however, increasing dimensions gives increasingly wrong timescales for tICA. Especially so when compared to $k$-means (triangle), PCA and the auto encoder which are not as strongly affected and are more stable over varying dimensionalities. These results are consistent with \cite{blochliger_weighted_2015} which shows that tICA is prone to larger errors when increasing dimensionality than other methods.

\subsection{Automated dimensionality detection}
\begin{figure}[]
\subfloat[PCA]{
  \includegraphics[width=.5\columnwidth]{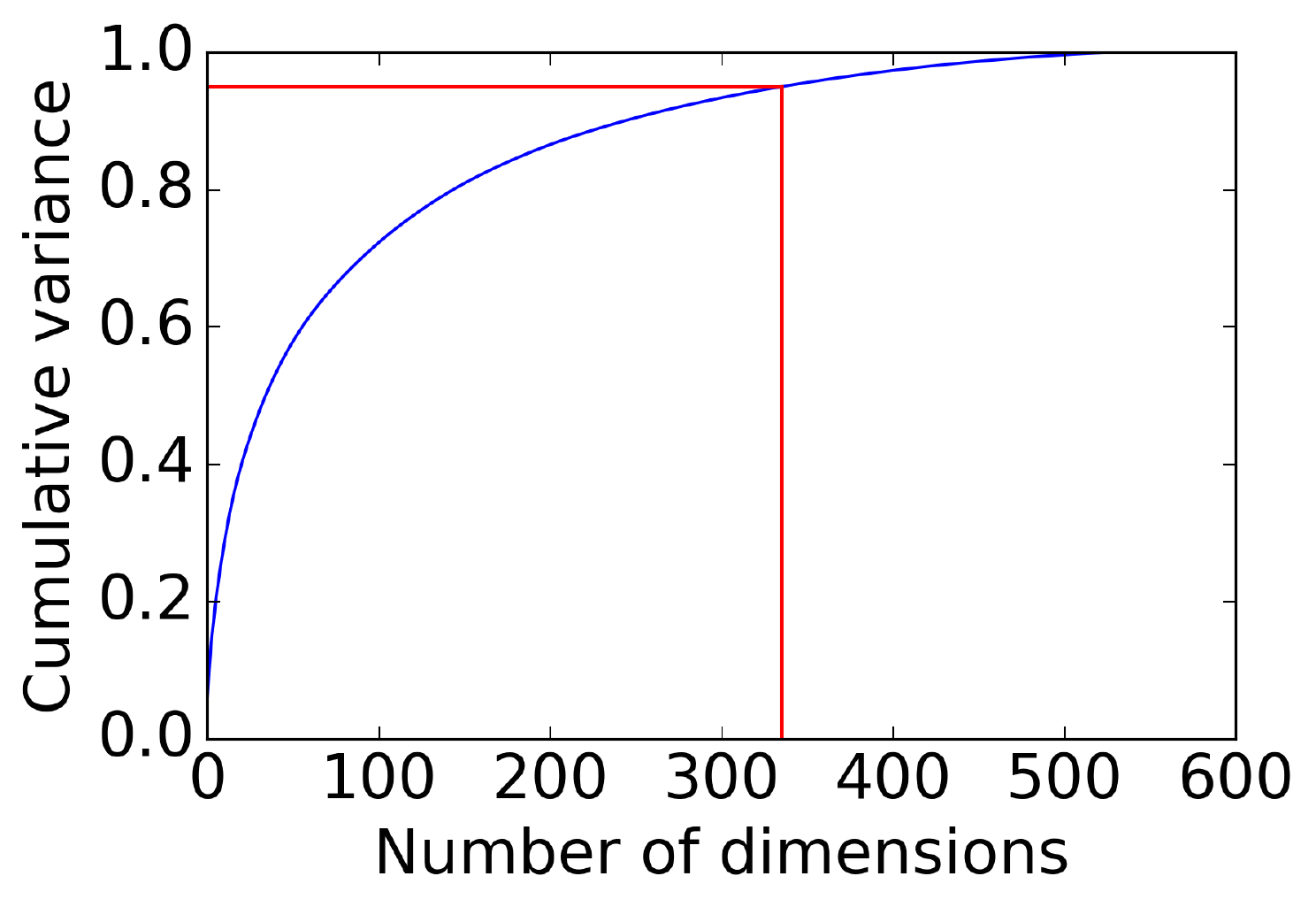}
}
\subfloat[tICA]{
  \includegraphics[width=.5\columnwidth]{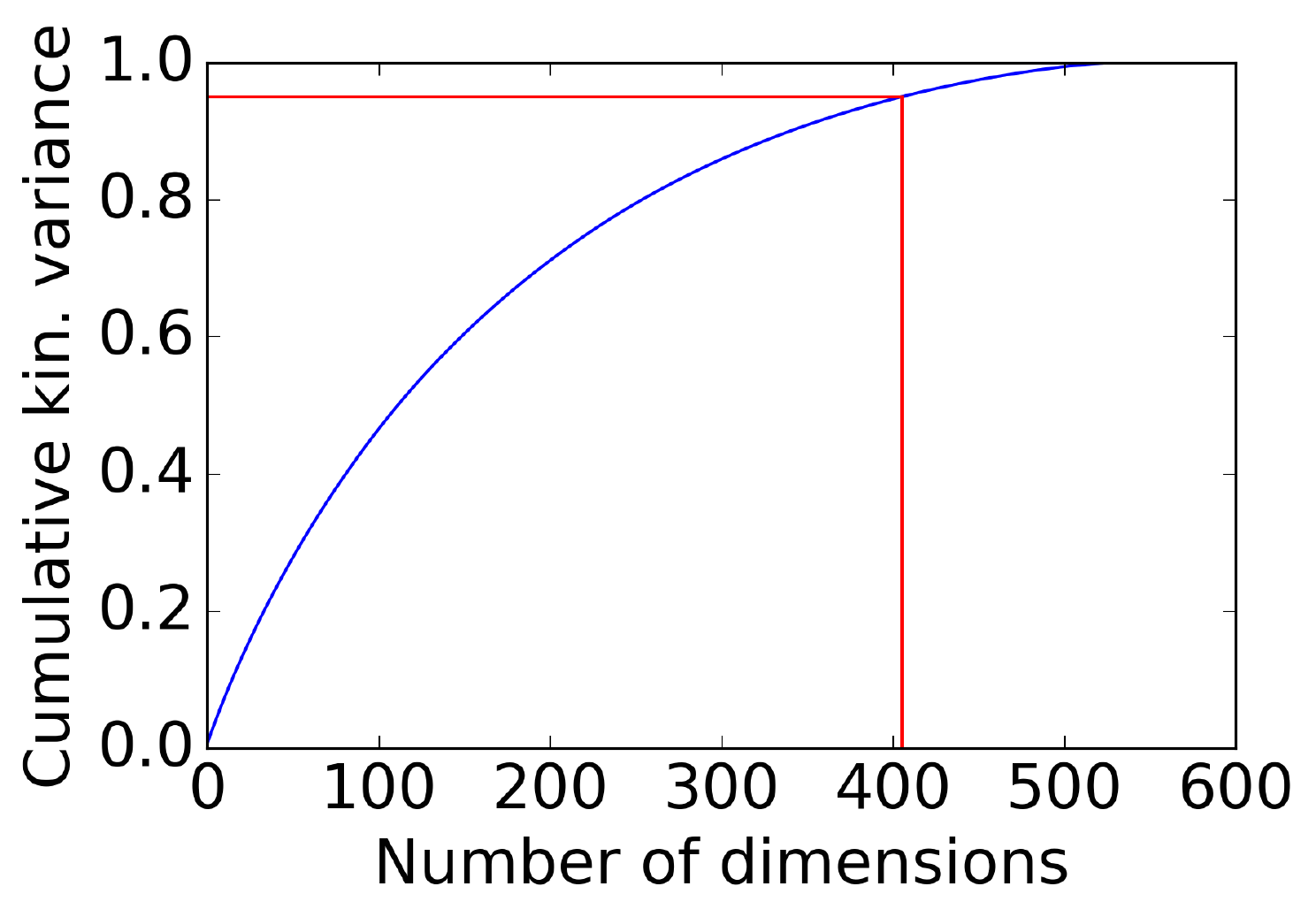}
}
\caption{Cumulative variance encoded in the PCA principal components and tICA independent components on the Villin data-set. Red lines show the 95\% variance cutoff.}
\label{fig:variancedim}
\end{figure}

As mentioned before, clustering methods have a hard time detecting clusters in high-dimensional spaces and the example of Villin consist further proof. A problem that arises from this, is the detection of the number of dimensions on which to project. As it strongly depends on the data-set in use, an automated method would be ideal to avoid having the user manually test multiple dimensions. Methods such as PCA and tICA, are able to calculate the percentage of variance described by the first $N$ principal and independent components and this is often used to calculate the ideal number of dimensions on which to project. Therefore, the problem of number of dimensions can be reinterpreted as deciding a specific variance percentage to keep when projecting. Typical heuristics include using the first $N$ dimensions which contain 95\% of the variance. However, as can be seen in Figure \ref{fig:variancedim}, this would produce a large number of dimensions (between 300 and 400 dimensions) for both PCA and tICA, failing to produce a functioning Markov model. Therefore, to our knowledge, there is currently no automated method for dimensionality detection that would be able to produce a functioning Markov model for the Villin data-set.

\subsection{Learned structural features}
\begin{figure}[!ht]
\input{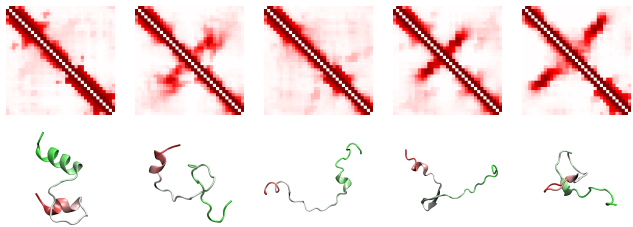}
\caption{
$k$-means (tri) cluster center contact maps on first row and centroid structures on second row.
}
\label{fig:vis_ktri}
\end{figure}

\begin{figure}[!ht]
\input{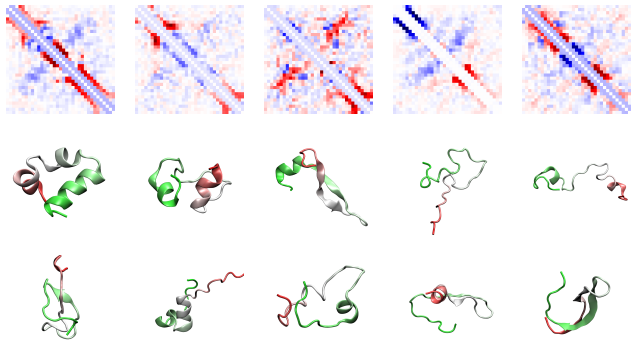}
\caption{
Auto encoder weights on first row and structures which maximally and minimally activate each hidden layer neuron on second and third row.
}
\label{fig:vis_ae}
\end{figure}

\begin{figure}[!ht]
\input{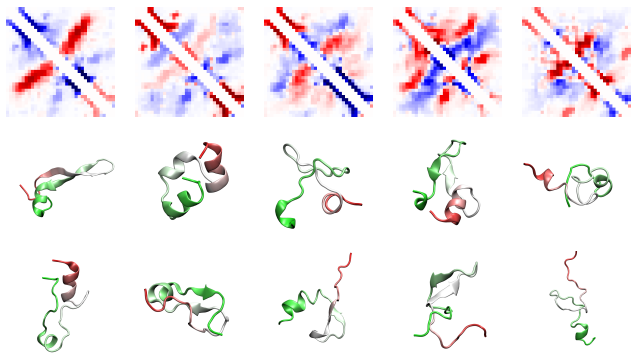}
\caption{
PCA principal components on first row and structures which correspond to the maximum and minimum values in each PC on second and third row.
}
\label{fig:vis_pca}
\end{figure}

\begin{figure}[!ht]
\input{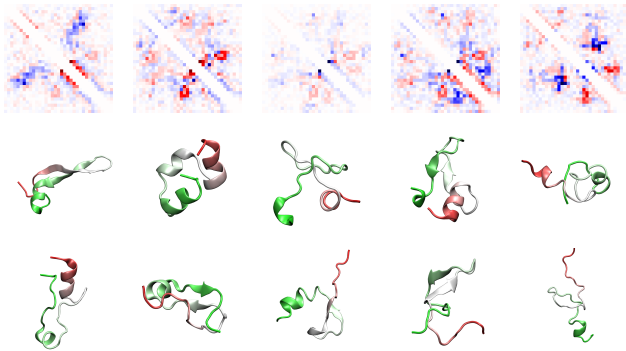}
\caption{
tICA independent components on first row and structures which correspond to the maximum and minimum values in each IC on second and third row.
}
\label{fig:vis_tica}
\end{figure}

To understand better how the dimensionality reduction methods work, it is interesting to visualize the features that the dimensionality reduction methods have learned. As we use protein contact maps for Villin, we are able to represent the weights, principal components, cluster centers and independent components respectively as two-dimensional images. We can see the different features that were learned by $k$-means (triangle) (Figure \ref{fig:vis_ktri}), the auto encoder (Figure \ref{fig:vis_ae}), PCA (Figure \ref{fig:vis_pca}) and tICA (Figure \ref{fig:vis_tica}). Red is used for positive weights, white for values close to zero and blue for negative weights. Under the weight maps, we show the protein conformations that most strongly represent those maps. For $k$-means (tri) we show the conformation of the centroids of the clusters. For the other methods, as they have negative and positive weights, we show in each column the two conformations that correspond to the maximum and minimum value along that dimension. All methods learned interesting features, both local and global. Features with strong positive or negative values close to the matrix diagonal encode local protein secondary structures; in this case alpha helices. On the other hand, features further from the diagonal encode more distant (global) residue interactions, and features perpendicular to the diagonal encode for anti-parallel beta strands which are a relatively common occurrence during simulations of Villin. The fully folded conformation of Villin consists of 3 folded helices (see Fig. \ref{fig:folded}a), therefore the folded conformation is represented by contact maps similar to the 1st cluster center in Figure \ref{fig:vis_ktri}, the first auto encoder hidden unit in Figure \ref{fig:vis_ae}, the second principal component in Figure \ref{fig:vis_pca} and the second independent component in Figure \ref{fig:vis_tica}.

\section{Conclusions}
In this paper we have used four methods for dimensionality reduction over two high-dimensional data-sets of protein folding and ligand binding trajectories. Benzamidine-Trypsin proved to be a trivial case in which dimensionality reduction is not necessary but can help improve the timescales for larger lag-times. On the other hand Villin proved much more complicated, where building a Markov model on the pure high-dimensional data was not able to reliably separate the underlying states to produce the correct kinetic quantities. However, a shallow auto encoder, a modified $k$-means featurization method, PCA and tICA were capable of improving the Markov model significantly. TICA, PCA and the auto encoder provided the best performance, however tICA proved to be very sensitive to the number of used dimensions. Indeed all methods were affected by the increase of dimensionality, albeit much less than tICA, indicating that the number of dimensions can be more important for MSM construction than the choice of projection method.

Both tICA and the auto encoder perform well on Villin using very low-dimensional projections, however, this could hide some slow dynamics in more complicated systems. An important remaining problem is the choice of dimensionality. Heuristics used for automatically detecting the best number of dimensions in PCA and tICA, such as the variance encoded in the first components fail to give good results in the case of a Markov model analysis. Therefore, it remains an open question as to how best detect the dimensionality required to analyze the system and currently manual testing needs to be done for each system to determine the dimensionality that gives the most consistent results.

Another factor that should to be taken into account when comparing projection methods is their run-time, as analysis of simulations can become computationally expensive for large data-sets. In this aspect $k$-means (triangle) is faster than the other methods, projecting the full Villin data-set on 10 dimensions in tens of seconds compared to tICA taking around 1 minute to calculate the ICs and projections, PCA 2.5 minutes and auto encoders on GPUs around 40 minutes to train and project the data.

As the first application of auto encoders on the dimensionality reduction of contact map data, we believe that they provided interesting results, even reaching the performance of other established methods. However, we believe there is more potential than is demonstrated here, as we were not yet able to exhaustively test different configurations of the networks. The large number of options in the construction of an auto encoder as well as the number of free parameters, allow for a variety of tuning and different setups. As auto encoders can go beyond the linearity of the other methods, we believe that they can show increased potential in other configurations and deeper architectures. Additionally, as auto encoders can learn local structural features they can become generalized and could potentially be applied to different data-sets of the same type (i.e. folding of different proteins) without the need of retraining.


\section{Appendix}
\begin{multline} \label{eq:ae}
J(W,b) = \bigg[ \frac{1}{m}\sum_{i=1}^{m}\Big(\frac{1}{2}\norm{\hat{x}_i - x_i}^2\Big)\bigg] + \\
\frac{\lambda}{2}\sum_{l=1}^{2}\sum_{i=1}^{s_l}\sum_{j=1}^{s_l+1}\Big(W_{ij}^{(l)}\Big)^2 + \\
\beta \sum_{j=1}^{s_2}KL(p||\hat{p}_j)
\end{multline}
where $m$ the number of training examples, $x_i$ training example $i$, $\hat{x}_i$ the reconstruction
of $x_i$ in the last layer of the auto encoder, $\lambda$ the weight decay parameter, $s_l$ the number of units in layer $l$,
$W^{(l)}$ the weight matrix of layer $l$, $\beta$ the sparsity penalty weight and $KL(p||\hat{p}_j)$ the Kullback-Leibler (KL) divergence
between $p$ the desired sparsity of the hidden units and $\hat{p}_j$ the mean sparsity of hidden unit $j$ over all training data.
In our setup we used $\lambda=0.003$, $p=0$ and $\beta=3$.

\begin{acknowledgement}
GDF aknowledges support from MINECO (BIO2014-53095-P) and FEDER. 
We thank the volunteers of GPUGRID for donating their computing time.
\end{acknowledgement}

\begin{suppinfo}
\end{suppinfo}

\bibliography{bibliography}

\end{document}